\documentclass[a4,a4wide]{article}
\usepackage{aaai}
\usepackage{times}
\usepackage{helvet}
\usepackage{courier}
\usepackage{url}
\usepackage{balance}
\begin{document}
 \nocopyright
%
\title{Some thoughts about benchmarks for NMR\thanks{This work has been supported in part by ANR Tuples.}}
\author{Daniel Le Berre\\
CNRS - Universit\'e d'Artois - France
}
\maketitle
\begin{abstract}
\begin{quote}
The NMR community would like to build a repository of benchmarks to push forward the design of systems implementing NMR as it has been the case for many other areas in AI.
There are a number of lessons which can be learned from the experience of other communities.
Here are a few thoughts about the requirements and choices to make before building such a repository.
\end{quote}
\end{abstract}

\section{What to expect}

Over the last two decades, a huge number of communities have built repositories of benchmarks, mainly with the idea to evaluate running systems on a common set of problems.
The oldest common input format for AI benchmarks is probably STRIPS \cite{STRIPS}, for planning systems.
One of the oldest and most compelling one for reasoning engines is TPTP (``Thousands of Problems for Theorem Provers'') \cite{TPTP}, the benchmarks library for First Order and Higher Order theorem provers. Such repository was built in 1993 and evolved since then as a companion to the CADE ATP System Competition (CASC) \cite{CASC}. There is an interplay between TPTP and CASC: TPTP is used to select benchmarks for CASC, benchmarks submitted to CASC are added eventually to TPTP and the solvers submitted to CASC are run on all TPTP benchmarks, and used to evaluate the practical complexity of those benchmarks. As such, over the years, benchmarks are ranked from hard to medium to easy with the improvements of the solvers. This is exactly the kind of virtuous circle one would like to see in each community. In the NMR community, a similar library exists with Asparagus\footnote{\url{http://asparagus.cs.uni-potsdam.de}}, which feeds the ASP competition \cite{ASPCOMP}.

There are however reasons which prevent it. Take for instance the SAT community. Its common input format is based on the Second Dimacs Challenge input format \cite{DIMACS}, one of the first SAT competitions. The benchmarks used for that competitive event has been a {\em de facto} standard for evaluating SAT solvers in practice. A system similar to TPTP was built by Laurent Simon in 2000:  SatEx \cite{SatEx}. However, the number of SAT solvers available in the SAT community became quickly much larger than the number of ATP systems, because of its increasing practical interest in hardware verification, and because it is much easier to develop a SAT solver than a First Order theorem solver. As such, it became quickly impossible to run all SAT solvers on all available benchmarks. A tradeoff was to organize a yearly SAT competitive event since 2002 \cite{SATCOMP2002}, to give a snapshot of the performances of recent solvers on a selection of benchmarks. 

\section{Modeling versus Benchmarking}

One of the first question which arises when creating a benchmark format is to be clear about the target of the format.
There are mainly two choices: one is to please the end user, by providing a format which simplifies modeling problems
in that format, the other one is to please the solver designers, to make sure that they integrate a way to read that format.
High level input format such as PDDL, TPTP, ASP, SMT and Minizinc (CSP) are clearly modeling oriented. Formats designed by the SAT community
(SAT, MAXSAT, PBO, QBF, MUS ...) are clearly solver oriented.

There are advantages and inconveniences for both approaches. The user oriented format favors the submissions of problems
by the community, because the input format is human understandable and easy to modify. However, such format may require a huge effort from the solver designer to adapt his solver to such format. This happened for instance for the SMT LIB 2 format, which was quite different from the original SMT LIB format, so it took time to be adopted by the SMT solver designers.
Another issue with user oriented formats are the potential high learning curve to understand all its subtleties. For instance, it took several
rounds in the Mancoosi International Solver Competition (MiSC) \cite{MISC} to see all solvers answering correctly to the requests because the input format 
was assuming some domain knowledge not obvious for a solver designer.

The main advantage of the solver oriented format is to be easy to integrate into any exiting system. It is the way to go if the community wants to evaluate
existing systems on a common basis. It was the idea behind the XCSP format for CSP solvers for instance \cite{XCSP}.
The major drawback of such approach is to force the end user to rely on an intermediate representation to generate those benchmarks,
and to perform some tasks by hand which may be automated using a higher level input format. For instance, in the case of SAT, it is required to translate the original
problem into propositional variables and clauses. Many users are not aware of basic principles and advanced techniques to perform those tasks efficiently.

One way to please both part is to provide a end-user input format, to favor the contribution of problems, and a solver input format to please the solver designers, with a default translator from the first one to the second one. This is the spirit of the Minizinc and Flatzinc formats in the CSP community \cite{MINIZINC}.

\section{Data versus Protocol}

Another question raised when designing an input format is whether the benchmark represents data or whether it represents a full protocol. The problem is orthogonal to the abstraction level of the input format: it is directed by the nature of the problems to be solved.

In many cases, benchmarks represent data, in one or multiple files (e.g. rules and facts, domain and instance), and the system answers to a single query. There are other cases in which some interaction with the system is required: the SMT LIB 2 format \cite{SMTLIB2} for instance defines a protocol to communicate with the system to solve problems incrementally, which means that the system in that case is stateful. The Aiger format used in the hardware model checking competition \cite{HWMCC} also provides some incremental capabilities, which corresponds to the unrolling of the Bounded Model Checking approach.

The protocol point of view is great for playing with toy examples, thus good for education. It also allows to interface with the solver without worrying about the details. From a system designer, it requires generally more effort to maintain the state of the system between queries.
From an efficiency point of view, an API is usually preferred in practice for interacting with a system.
\section{Checkable queries}

Once a common benchmark format is setup, it is important to make sure that the benchmarks 
are correctly read by the systems, and that the queries to the systems provide answers checkable by a third party tool.
In the case of SAT for instance, while the decision problem answer is yes or no, in practice, the SAT solvers have always been asked to provide a certificate (a model) in case of satisfiability. Such certificate can be checked by an independent tool: if it satisfies all clauses, then the answer is checked, else the answer is invalid.
If two solvers disagreed on the satisfiability of a benchmark, checking the certificate of the yes answer allowed to spot incorrect solvers when that certificate was correct: the no answer is clearly incorrect in that case.
Nothing could be decided if the certificate was invalid: there are many reasons why a SAT solvers could answer SAT and provide an incorrect certificate (complex pre-processing and in-processing being the most probable case). 
There has been since 2005 an effort to also provide checkable no answers to SAT solvers  \cite{CERTUNSATALLEN}, but very few solver designers implemented it until a simpler 
proof certificate requiring to add only a few lines of code in the solver was designed in 2013  \cite{CERTUNSATMARIJN}. As such, SAT solvers answers can now be checked both in case of satisfiability and unsatisfiability.

Note that it is not always possible to check the system answer. It happens for instance for QBF solvers, for which a certificate would be a winning strategy for the existential player.
During the QBF evaluations, many QBF solvers disagreed on the status of the benchmarks. As such, several approaches were taken to sort out the situation: majority voting, let the solvers play against each other \cite{CERTQBFARMANDO}, fuzz testing and delta debugging \cite{CERTQBFARMIN}.
It also happens when computing an optimal solutions in Pseudo-Boolean Optimization or MaxSat competitions: in that case, one just check the value of the certificate returned by the solver, and that no other solver found a better solution. A better but resource consuming approach would be to create a new benchmark to check that there is no better solution. In the same spirit, when tools for computing Minimal Unsatisfiable Subformula are used, it is very demanding to check for each answer that 
both the set of constraints is unsatisfiable and that removing any clause makes the set of constraints satisfiable. In the MUS track of the SAT competition 2011, only the first test was performed, offline.

It is important in the first place to provide both to the end users and the solver designers some sample benchmarks with their expected answer, or a basic solver able to solve small benchmarks. This is especially true if the input format is user oriented.
For instance, the MISC competition  introduced new features in the input format without providing sample benchmarks with those new features.
Those features were not correctly implemented by all systems, thus the systems answered differently on some of the benchmarks, making comparisons between the systems hardly possible. 

\section{Chicken and egg problem}

It is unlikely that people start providing benchmarks in one input format without having a system to test some reduced scale benchmarks.
It is also unlikely that solver designers start supporting an input format without having some sample benchmarks to play with. That's the reason why 
a common input format is a community effort and it relies generally on a small group of people who are concerned by the subject.
One can take as example the attempt during the SAT 2005 competition to push forward a non CNF input format for SAT\footnote{\url{http://www.satcompetition.org/2005/}}: a common input format was defined, allowing to define arbitrary gates, and a few sample instances were provided as part of a specific track of the competition.  No submission of benchmarks nor systems were received for such track. Another attempt, using a more
specific non clausal format (And Inverter Graph, AIG), but well suited for model checking, received more interest in 2007, and became a competition on its own for hardware model checking \cite{HWMCC}. The main difference between the two attempts was that a small community agreed to support AIG, some translators and checkers were available (AIGER tool suite\footnote{\url{http://fmv.jku.at/aiger/}}) and many model checking benchmarks were provided in such format.

The input format of a given system may become a {\em de facto} common input format. In the case of argumentation frameworks for instance, several systems based on different technologies have been designed by the same group, using a common input format\footnote{\url{http://www.dbai.tuwien.ac.at/research/project/argumentation/}}. Such input format could be a good starting point for creating a common argumentation system input framework.

If it is not possible to provide both some sample benchmarks and a basic solver, it is important to provide a way to check the answers. The minimum requirement here would be to provide the expected answer for each sample benchmark in a text file.
A better approach would be to provide a way to check the answer thanks to a certificate using an independent checker software. Note that in such a case,  a common output (certificate) format must also be defined. 

\section{Reusing benchmarks from other communities}

Reusing benchmarks from other communities is certainly an easy way to start collecting benchmarks. Most benchmarks libraries contain well-known academic benchmarks (including randomly generated ones), benchmarks based on other community benchmarks (SAT has many benchmarks modeling properties to check on circuit benchmarks from ISCAS for instance), and finally dedicated benchmarks.
The latter are the harder to find at the beginning.  As such, reusing benchmarks from other communities is often the only way to retrieve non-academic benchmarks.

Note that there are some side effects in reusing benchmarks from other communities.
The first one is to pay attention when evaluating systems on the origin of those systems.
For instance, there are two optimization extensions to SAT for which benchmarks are available: MAXSAT and Pseudo Boolean Optimization. The PBO benchmarks appeared before the MAXSAT ones, and some benchmarks from PBO have been expressed as MAXSAT problems (optimization problems with one linear objective function and a set of clauses can be equally expressed in both frameworks). Some solvers designed to solve PBO problems have been extended to solve MAXSAT problems (e.g. Sat4j). Those solvers usually perform very well on the benchmarks originating from PBO.
In the same spirit, some of the Pseudo Boolean benchmarks are coming from MIPLIB\footnote{\url{http://miplib.zib.de}}, a repository of Mixed Integer Linear Programming benchmarks used by MILP optimizers developers since 1992 to evaluate their systems. It is no surprise if tools such as CPLEX performs very well on those benchmarks when compared to ``classical'' Pseudo-Boolean solvers.

In the case of NMR, it is often the case that the systems have to deal with inconsistency. As such, it is tempting for instance to use unsatisfiable SAT benchmarks to evaluate NMR systems. But those systems usually require additional informations (e.g. a stratification of the clauses, a confidence for each clause, etc) and some arbitrary choices would have to be done to fit in the context (i.e. creating individual satisfiable sub-CNF for each agent in a multi-agent context). The additional information may be generated using a specific distribution of values (e.g. randomly and uniformly assigning the clauses to a given number of strata), or arbitrarily (e.g. make strata from sets of  consecutive clauses, of identical or random sizes). Those benchmarks, despite not being related at all with a real NMR problem, do have the benefit to allow different systems to be compared on the same basis.

It is also interesting to note that there exists a format in the SAT community which is very close to stratified knowledge bases: {\em Group oriented CNF},  introduced in the MUS special track in the SAT 2011 competition \footnote{\url{http://www.satcompetition.org/2011/rules.pdf}}. The benchmarks in that format are coming from circuit designs \cite{HLMUS1,HLMUS2}, where each group (stratum) of clauses correspond to a subcircuit, a specific group contains hard clauses which correspond to integrity constraints  (i.e. knowledge) while the remaining groups are soft clauses which can be enabled or disabled altogether (i.e. beliefs). The benchmarks are not satisfiable if all groups of clauses are enabled. There exists 197 group oriented CNF benchmarks available from the SAT 2011 competition web site, all corresponding to ``real" designs. They could be a good starting point to test systems requiring stratified knowledge bases. 

\section{The bias of benchmarking systems}

It should also be clear that the benchmarks used to evaluate the systems drive in some sense which systems are going to be developed or improved by the community.

Anyone looking at the winners of the various SAT competitions\footnote{\url{http://www.satcompetition.org/}} can check that solvers behave differently on randomly generated benchmarks and benchmarks coming from real applications or hard combinatorial problems. This is true for any community. Randomly generated benchmarks are interesting for two reasons: they are easy to generate and can generally be formally defined. Combinatorial benchmarks are important because they usually force the system to exhibit worst case behavior. Application benchmarks are interesting because they provide some hints about the practical complexity of the problem. Note that if application benchmarks in SAT tend to be ``easier'' in practice than say combinatorial benchmarks, it is only the case because people worked hard to find the right heuristics, data structures, etc. to manage those problems.

For that reason, one should always be very careful when looking at any competitive event results, or when evaluating his system on a given set of benchmarks.
It took some time for the MAXSAT competition\footnote{\url{http://maxsat.ia.udl.cat/}} to obtain benchmarks coming from real applications. Before 2008, SAT-based MAXSAT solvers performed relatively poorly on the problems available for the competition (mainly randomly generated, based on academic problems). Once application benchmarks became available, SAT-based MAXSAT solvers performed much better on those problems, especially core-guided MAXSAT solvers.
So the benchmarks used to evaluate the systems eventually influence the development of those systems.

There are also subtle differences between benchmarks coming from real applications.
The SAT community has been driven by Bounded Model Checking benchmarks from the end of the 90's to mid 2000's. As such, the solvers designed during that period were especially relevant to that application: the winners of the SAT competition could be directly integrated into model checkers. With an increase of the diversity of its applications, the available benchmarks for SAT are now quite different in structure from those BMC benchmarks. Which means that the best performing SAT solver during the SAT competition may not be the best solver for the particular case of BMC.

\section{Benchmarks libraries}

Benchmarks are usually made available to the community through a library: CSPLIB, SATLIB, PBLIB, SMTLIB, etc. However, it is an issue to manage those libraries in the long term.
A good example is SATLIB \cite{SATLIB}. It was designed in 1999 to host the benchmarks made available to the SAT community. It did a good job at collecting the benchmarks generated during the 90's. However, the huge increase in number of benchmarks (and their size!) in early 2000 made it hard to catch up after 2001, so the SAT competition web sites have been providing the benchmarks used in the competitions since then. The situation is not ideal because there is no longer now in the SAT community a central place where the benchmarks can be accessed.
Some of the benchmarks, which were made available to the research community by IBM \cite{ZARPAS}, can no longer be distributed. It is thus very difficult to reproduce some experiments, to evaluate the efficiency of new solvers on those benchmarks. Having a community driven central repository may help to avoid such situation.

The CSP library \footnote{\url{http://www.csplib.org/}} succeeded in maintaining a library of problems for 15 years. Note that those problems are not in a uniform format, but rather described in their own format. The library is much about problems than benchmarks.

The library of benchmarks one community would like to mimic today are probably TPTP\footnote{\url{http://www.tptp.org/}} or MIPLIB. Those libraries have been available for two decades now and are the central sources of benchmarks for their respective community.
The benchmarks are ranked by difficulty, and updated regularly at the light of the performances of new systems. 

\section{Conclusion}

Many communities built central repositories of benchmarks to be able to compare the performance of their systems. The success of those repositories relies first on the adoption of it format by the community, and second on the availability of benchmarks for which some information is provided: difficulty, expected answer, runtime of existing systems, etc. 

For a community such as NMR, which addresses a wide range of different problems, the first step is to decide on which problems a first effort of standardization is required. The heuristics can be either the maturity of existing systems in the community or the importance of the problem for the community. In either case, the choice of the format for the benchmarks will be important: should it be user oriented or system oriented? data or protocol oriented?

Defining a format and providing benchmarks is not sufficient to reach adoption: sample results and answers checkers are essential components to allow system designers to adopt such format.
In order to receive application benchmarks, some systems supporting that format should be provided as well, even if they are not very efficient: they are sufficient to discover the meaning of the benchmark format, or to check the answers of a system under development.

Both benchmarks providers and system developers can make mistakes. As such, tools which check the syntax of the input and the correctness of the system answers will help providing meaningful benchmarks and systems results.

In order to reuse benchmarks from other communities, tools which allow to translate to and from different formats are also welcome. 

Organizing competitive events has been a great source of new benchmarks for many communities.
I am looking forward the first NMR competition.
\bibliographystyle{aaai}
\bibliography{nmr14}

\begin{thebibliography}{}

\bibitem[\protect\citeauthoryear{Abate and Treinen}{2011}]{MISC}
Abate, P., and Treinen, R.
\newblock 2011.
\newblock {Mancoosi Deliverable D5.4: Report on the international competition}.
\newblock Rapport de recherche.

\bibitem[\protect\citeauthoryear{Barrett, Stump, and Tinelli}{2010}]{SMTLIB2}
Barrett, C.; Stump, A.; and Tinelli, C.
\newblock 2010.
\newblock {The SMT-LIB Standard: Version 2.0}.
\newblock In Gupta, A., and Kroening, D., eds., {\em Proceedings of the 8th
  International Workshop on Satisfiability Modulo Theories (Edinburgh, UK)}.

\bibitem[\protect\citeauthoryear{Biere and Jussila}{2007}]{HWMCC}
Biere, A., and Jussila, T.
\newblock 2007.
\newblock Hardware model checking competition.
\newblock http://fmv.jku.at/hwmcc07/.

\bibitem[\protect\citeauthoryear{Brummayer, Lonsing, and
  Biere}{2010}]{CERTQBFARMIN}
Brummayer, R.; Lonsing, F.; and Biere, A.
\newblock 2010.
\newblock Automated testing and debugging of sat and qbf solvers.
\newblock In Strichman, O., and Szeider, S., eds., {\em SAT}, volume 6175 of
  {\em Lecture Notes in Computer Science},  44--57.
\newblock Springer.

\bibitem[\protect\citeauthoryear{Fikes and Nilsson}{1971}]{STRIPS}
Fikes, R., and Nilsson, N.~J.
\newblock 1971.
\newblock Strips: A new approach to the application of theorem proving to
  problem solving.
\newblock {\em Artif. Intell.} 2(3/4):189--208.

\bibitem[\protect\citeauthoryear{Gebser \bgroup et al\mbox.\egroup
  }{2007}]{ASPCOMP}
Gebser, M.; Liu, L.; Namasivayam, G.; Neumann, A.; Schaub, T.; and
  Truszczynski, M.
\newblock 2007.
\newblock The first answer set programming system competition.
\newblock In Baral, C.; Brewka, G.; and Schlipf, J.~S., eds., {\em LPNMR},
  volume 4483 of {\em Lecture Notes in Computer Science},  3--17.
\newblock Springer.

\bibitem[\protect\citeauthoryear{Heule, Jr., and
  Wetzler}{2013}]{CERTUNSATMARIJN}
Heule, M.; Jr., W. A.~H.; and Wetzler, N.
\newblock 2013.
\newblock Verifying refutations with extended resolution.
\newblock In Bonacina, M.~P., ed., {\em CADE}, volume 7898 of {\em Lecture
  Notes in Computer Science},  345--359.
\newblock Springer.

\bibitem[\protect\citeauthoryear{Hoos and Stützle}{2000}]{SATLIB}
Hoos, H.~H., and Stützle, T.
\newblock 2000.
\newblock Satlib: An online resource for research on sat.
\newblock In Gent, I.~P.; van Maaren, H.; and Walsh, T., eds., {\em SAT 2000},
  283--292.
\newblock IOS Press.

\bibitem[\protect\citeauthoryear{Johnson and Trick}{1996}]{DIMACS}
Johnson, D., and Trick, M., eds.
\newblock 1996.
\newblock {\em Second {DIMACS} implementation challenge : cliques, coloring and
  satisfiability}, volume~26 of {\em DIMACS Series in Discrete Mathematics and
  Theoretical Computer Science}.
\newblock American Mathematical Society.

\bibitem[\protect\citeauthoryear{Lecoutre, Roussel, and van
  Dongen}{2010}]{XCSP}
Lecoutre, C.; Roussel, O.; and van Dongen, M. R.~C.
\newblock 2010.
\newblock Promoting robust black-box solvers through competitions.
\newblock {\em Constraints} 15(3):317--326.

\bibitem[\protect\citeauthoryear{Nadel}{2010}]{HLMUS1}
Nadel, A.
\newblock 2010.
\newblock Boosting minimal unsatisfiable core extraction.
\newblock In Bloem, R., and Sharygina, N., eds., {\em FMCAD},  221--229.
\newblock IEEE.

\bibitem[\protect\citeauthoryear{Narizzano \bgroup et al\mbox.\egroup
  }{2009}]{CERTQBFARMANDO}
Narizzano, M.; Peschiera, C.; Pulina, L.; and Tacchella, A.
\newblock 2009.
\newblock Evaluating and certifying qbfs: A comparison of state-of-the-art
  tools.
\newblock {\em AI Commun.} 22(4):191--210.

\bibitem[\protect\citeauthoryear{Ryvchin and Strichman}{2011}]{HLMUS2}
Ryvchin, V., and Strichman, O.
\newblock 2011.
\newblock Faster extraction of high-level minimal unsatisfiable cores.
\newblock In Sakallah, K.~A., and Simon, L., eds., {\em SAT}, volume 6695 of
  {\em Lecture Notes in Computer Science},  174--187.
\newblock Springer.

\bibitem[\protect\citeauthoryear{Simon and Chatalic}{2001}]{SatEx}
Simon, L., and Chatalic, P.
\newblock 2001.
\newblock Satex: A web-based framework for sat experimentation.
\newblock {\em Electronic Notes in Discrete Mathematics} 9:129--149.

\bibitem[\protect\citeauthoryear{Simon, {Le Berre}, and
  Hirsch}{2005}]{SATCOMP2002}
Simon, L.; {Le Berre}, D.; and Hirsch, E.~A.
\newblock 2005.
\newblock {The SAT2002 competition}.
\newblock {\em Ann. Math. Artif. Intell.} 43(1):307--342.

\bibitem[\protect\citeauthoryear{Stuckey, Becket, and Fischer}{2010}]{MINIZINC}
Stuckey, P.~J.; Becket, R.; and Fischer, J.
\newblock 2010.
\newblock Philosophy of the minizinc challenge.
\newblock {\em Constraints} 15(3):307--316.

\bibitem[\protect\citeauthoryear{Sutcliffe and Suttner}{2006}]{CASC}
Sutcliffe, G., and Suttner, C.
\newblock 2006.
\newblock {The State of CASC}.
\newblock {\em AI Communications} 19(1):35--48.

\bibitem[\protect\citeauthoryear{Sutcliffe}{2009}]{TPTP}
Sutcliffe, G.
\newblock 2009.
\newblock {The TPTP Problem Library and Associated Infrastructure: The FOF and
  CNF Parts, v3.5.0}.
\newblock {\em Journal of Automated Reasoning} 43(4):337--362.

\bibitem[\protect\citeauthoryear{{Van Gelder}}{2012}]{CERTUNSATALLEN}
{Van Gelder}, A.
\newblock 2012.
\newblock Producing and verifying extremely large propositional refutations -
  have your cake and eat it too.
\newblock {\em Ann. Math. Artif. Intell.} 65(4):329--372.

\bibitem[\protect\citeauthoryear{Zarpas}{2006}]{ZARPAS}
Zarpas, E.
\newblock 2006.
\newblock {Back to the SAT05 Competition: an a Posteriori Analysis of Solver
  Performance on Industrial Benchmarks}.
\newblock {\em JSAT} 2(1-4):229--237.

\end{thebibliography}
\end{document}